\documentclass[pdflatex,sn-mathphys-num]{sn-jnl}

\usepackage{graphicx}%
\usepackage{multirow}%
\usepackage{amsmath,amssymb,amsfonts}%
\usepackage{amsthm}%
\usepackage{mathrsfs}%
\usepackage[title]{appendix}%
\usepackage{xcolor}%
\usepackage{textcomp}%
\usepackage{manyfoot}%
\usepackage{booktabs}%
\usepackage{algorithm}%
\usepackage{algorithmicx}%
\usepackage{algpseudocode}%
\usepackage{listings}%

\usepackage{tcolorbox}
\tcbuselibrary{listings}

\theoremstyle{thmstyleone}%
\theoremstyle{thmstyletwo}%

\theoremstyle{thmstylethree}%
\newcommand{\copulasymbol}[1]{%
  \mathrel{\mathchoice
    {\mathbin{#1}}
    {\mathbin{#1}}
    {\mathbin{\scriptstyle #1}}
    {\mathbin{\scriptstyle #1}}
  }%
}

\newcommand{\predictiveImplication}{\copulasymbol{\ooalign{\kern0.22em$\vcenter{\hbox{\scalebox{0.6}[0.5]{\textbf{/}}}}$\hfill\cr$\Rightarrow$}}} 
  
\newcommand{\retrospectiveImplication} {\copulasymbol{\ooalign{\kern0.22em$\vcenter{\hbox{\scalebox{0.6}[0.5]{\textbf{\textbackslash}}}}$\hfill\cr$\Rightarrow$}}} 
\newcommand{\concurrentImplication}{\copulasymbol{\ooalign{\kern0.32em$\vcenter{\hbox{\scalebox{0.6}[0.5]
{\textbf{\textbar}}}}$\hfill\cr$\Rightarrow$}}} 

\begin{document}

\title[Hallucination is Inevitable]{Hallucination is Inevitable for LLMs with the Open World Assumption}


\author*{\fnm{Bowen} \sur{Xu}}\email{bowenxu.agi@gmail.com}\email{bowen.xu@temple.edu}


\affil{\orgdiv{Department of Computer and Information Sciences}, \orgname{Temple University}, \orgaddress{\street{1801 N Broad St}, \city{Philadelphia}, \postcode{19122}, \state{PA}, \country{USA}}}

\abstract{
Large Language Models (LLMs) exhibit impressive linguistic competence but also produce inaccurate or fabricated outputs, often called ``hallucinations''. Engineering approaches usually regard hallucination as a defect to be minimized, while formal analyses have argued for its theoretical inevitability. Yet both perspectives remain incomplete when considering the conditions required for artificial general intelligence (AGI).  
This paper reframes ``hallucination'' as a manifestation of the generalization problem. Under the \textit{Closed World assumption}, where training and test distributions are consistent, hallucinations may be mitigated. Under the \textit{Open World assumption}, however, where the environment is unbounded, hallucinations become inevitable. 
This paper further develops a classification of hallucination, distinguishing cases that may be corrected from those that appear unavoidable under open-world conditions. 
On this basis, it suggests that ``hallucination'' should be approached not merely as an engineering defect but as a structural feature to be tolerated and made compatible with human intelligence.
}


\maketitle

\section{Introduction}

Large Language Models (LLMs)~\cite{bubeck2023sparks-agi} have rapidly become central to contemporary artificial intelligence, influencing domains as diverse as information retrieval, scientific discovery, and everyday decision-making. Yet their increasing deployment has brought renewed attention to a longstanding difficulty: the phenomenon of ``hallucination''~\cite{huang2025survey-hallucination}. LLMs sometimes generate information that is fluent and persuasive but nonetheless inaccurate or fabricated. In engineering discourse, hallucination is typically treated as a defect to be minimized through better datasets, architectures, or alignment procedures~\cite{cossio2025taxonomy-hallucination, kalai2025why}. In philosophical discourse, however, the status of hallucination is far less straightforward: is it a contingent flaw of current systems, or is it in some sense inevitable for any learner operating in open-ended environments?

Recent research has attempted to formalize this issue. Some authors define hallucination as output error relative to a ground-truth and argue that such errors are theoretically unavoidable given infinite input spaces and adversarial environments.~\cite{xu2024hallucination} Some others take a probabilistic perspective, showing that while hallucinations cannot be eliminated in principle, they can be rendered statistically negligible with sufficient data and appropriate algorithms~\cite{suzuki2025hallucination}, and the analyses presuppose what might be called the ``Closed World assumption''\footnote{This usage should not be confused with the so-called Closed World Assumption in logic, which assumes that any statement not contained in a knowledge base is false.} termed herein, in which the training distribution is assumed to adequately represent the target environment. Under this assumption, hallucination appears controllable: additional data, improved inference, or cautious abstention (\textit{i.e.}, ``I don’t know'') may reduce error to low levels.

This paper takes a different stance and argues that when the ``Open World assumption'' -- a condition more faithful to real-world intelligence and to the ambitions of artificial general intelligence -- is adopted, hallucination becomes not merely difficult to avoid but inevitable. The Open World assumption maintains that intelligent systems must cope with unbounded space and time, and tasks. In such a setting, past experience does not guarantee future accuracy; generalization is both indispensable and fallible. Just as humans mis-generalize, or form inductive expectations that later prove false, LLMs cannot hope to avoid hallucination. Hallucination, in this light, should be seen not as a marginal bug but as a structural byproduct of intelligence under open-world conditions.

The contribution of this paper is threefold. First, the hallucination problem is reduced to the generalization problem in the traditional machine learning research Second, three types of errors are distinguished, and the latter two are viewed as hallucination. Type-I hallucination arises from false memorization: the model produces an answer inconsistent with facts already present in its training data. Such errors are, in principle, corrigible. Type-II hallucination, by contrast, arises from false generalization to cases not present in the training set. This latter form is inescapable under the Open World assumption, since no finite past can guarantee correctness in an infinite future. Third, constructive ways of ``treating'' hallucination are proposed: tolerating it as a normal phenomenon of deep learning mechanisms, designing systems that remain adaptive in the face of limited experience, and aligning error patterns with forms of reasoning that are intelligible and acceptable to humans.

\section{Debates on Inevitability of ``Hallucination''}

LLMs' behaviors of generating false information have been noticed in the early stage. The formal discussion on the inevitability of LLM hallucination can be traced back to \citet{xu2024hallucination}: the authors defined LLM hallucination as the output error compared with a ground-truth, and they argued that LLM hallucination is inevitable in the sense that there exists a function $f$, representing a world, which makes the sole ground-truth different from an LLM's output at a certain training stage, and any LLMs would make infinitely many errors in that world. 
Although this paper draws a similar conclusion as \cite{xu2024hallucination}, the meaning and line of argument are fundamentally different. 
I disagree with \citet{xu2024hallucination}'s justification in three aspects. 
First, I do not view arbitrary types of errors as ``hallucination''. Humans make mistakes due to many reasons. One origin is lacking information, as a result, they have beliefs that are not consistent with some facts. 
Additionally, some errors stem from the changes of outside standards. For example, an LLM can always generate wrong answers in an adversarial world (see Fig.~\ref{fig:adversarial-world} in Appendix~\ref{ap:counter-example}).
In this paper, as argued in Sec.~\ref{sec:generalize} and \ref{sec:inevitable},
``hallucination'' can be viewed as a byproduct of generalization: it can be reasonable to some extent, but the correctness cannot be guaranteed under the Open World assumption.
Second, proofs in \citet{xu2024hallucination} rely on an infinite set of samples (otherwise, the statement that an LLM makes infinitely many errors cannot be true). However, in practice, the lengths of strings input to and output from an LLM are always finite, which makes the sample space finite. Therefore, if collecting enough data, with enough resources, all samples are contained in the training-set, and errors can be fully avoided in theory. 
Third, \citet{xu2024hallucination} assume that each input corresponds to a single correct answer as the ground-truth, while any different answer from the ground-truth is viewed as error. However, due to the many-to-many mapping between sentences and meanings, this assumption is not suitable. At the worst case, an LLM can answer ``I don't know'', which is different from the ground-truth but should not be simply viewed as a wrong answer.
For these reasons, while I acknowledge the formal validity of their theorems within their chosen framework, I find their definitions and assumptions too restrictive for capturing the phenomenon of hallucination as it arises in practice.
My analysis therefore shifts the focus: from existence proofs to the epistemic and methodological inevitability of hallucination under open-world conditions.

\citet{suzuki2025hallucination} criticized \citet{xu2024hallucination} from a different perspective. \citet{suzuki2025hallucination} introduced the notion of hallucination probability to measure the likelihood that a language model generates outputs outside the acceptable set under a given input distribution. While computability theory shows that hallucinations are inevitable on an infinite set of inputs, the authors prove from a probabilistic perspective that the probability of hallucinations can be made arbitrarily close to zero, provided the training data is sufficiently high-quality and abundant, and appropriate training and inference algorithms are used.
In other words, hallucinations are theoretically unavoidable in principle, but in practice they can be rendered statistically negligible, according to\citet{suzuki2025hallucination}. Therefore, if hallucinations persisted in real applications, the cause should be attributed to limitations in the dataset or algorithms, rather than to any ``innate inevitability'' dictated by computability theory. I agree with \citet{suzuki2025hallucination} that hallucination can be largely mitigated when the test-set is fully consistent with the training-set (as termed herein ``the closed environment assumption''). However, the Open World assumption should be adopted for some reasons, which negate the guarantee of the consistency (see Sec.~\ref{sec:inevitable}).

Undoubtedly, current LLMs cannot answer any questions without errors. Some researchers summarized types of problems that LLMs cannot solve.~\cite{shi2025hallucination, banerjee2025hallucination} For example, some problems have no solution logically, some solutions are hard to compute, and some solutions suffers from resource limitation. 
Yet humans also cannot solve those problems for the same reasons.
In these three situations, hallucinations are still evitable -- LLMs can answer ``Any answer is wrong for this question'', ``I am not able to solve that'', ``I cannot do that'', \textit{etc}.
Many problems that cause LLMs' ``inevitable'' wrong answers are actually evitable by answering, \textit{e.g.}, ``I don't know''.

\section{Generalization and the ``No Free Lunch'' Theorem}\label{sec:generalize}

Beyond these debates, this paper advances a distinct perspective: hallucination should be understood as a manifestation of the generalization problem in machine learning.

The operation of an LLM can be abstracted at the level of single-token prediction. Concretely, each step can be represented as a mapping $x \mapsto g(x)$, where $x$ denotes the current context (a tensor) and $g(x)$ is the embedding of the next predicted token. This mapping is applied repeatedly to extend the sequence. For example, define $g_0(x)=g(x)$, $g_{1}(x)=g([x,g_0(x)])$, $g_{2}(x)=g([x,g_0(x),g_1(x)])$, and so on, up to $g_n(x)=g([x,g_0(x),g_1(x),...,g_{n-1}(x)])$. 
On this basis, the overall sequence-generation process can itself be viewed as a higher-level mapping,  
$f(x) = [g_0(x), g_1(x), \ldots, g_n(x), \diamond, \ldots, \diamond]$,
where ``$\diamond$'' denotes a placeholder (\textit{e.g.}, an all-zero tensor) to ensure a constant output dimension $n_{\text{max}}$, corresponding to the maximal number of tokens an LLM can produce in practice. Although LLMs are in principle capable of unbounded continuation, their outputs are always restricted to a finite length; the above abstraction adopts this restriction for theoretical clarity.

The mapping $f(\cdot)$ is learned from training samples through different mechanisms. For instance, $g(\cdot)$ can be trained by minimizing the prediction error between its output $\hat{g}(x)$ and the next token. In addition, $g(\cdot)$ -- and hence $f(\cdot)$ -- can be further adjusted by other mechanisms, such as \textit{Reinforcement Learning from Human Feedback} (RLHF)~\cite{kaufmann2024rlhf-survey}, where updates depend on the evaluation of output sequence $\hat{f}(x)$ against reward signals.

Therefore, an LLM can be abstracted as a mapping $x \mapsto f(x)$, where both $x$ and $f(x)$ are tensors. In this sense, a large language model may be regarded as a type of machine learning model. Although its training procedures differ from those of traditional classifiers and function approximators, many insights from classical machine learning -- particularly those concerning generalization -- remain applicable.

\begin{figure}[h]
    \centering
    \includegraphics[width=0.38\linewidth]{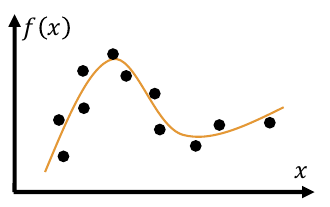}
    \caption{Illustration of a learned mapping. Black dots represent training samples, while the orange curve denotes the mapping acquired by an LLM. The reduction to two dimensions is for intuitive visualization; the underlying analysis applies equally to high-dimensional spaces.\protect\footnotemark}
    \label{fig:case1}
\end{figure}

For ease of exposition, this mapping can be illustrated in a two-dimensional diagram (Fig.~\ref{fig:case1}). The reduction to two dimensions is purely for intuitive visualization; the underlying analysis applies equally to the high-dimensional case. In the figure, black dots represent training samples, while the orange curve depicts the mapping learned by an LLM.

\footnotetext{Color versions of this and all subsequent figures are available in the online version.}

An LLM maps an input to an output. However, training sets typically do not cover all possible input–output relations. As a result, an LLM must generalize to novel situations based on its prior experience. Even when large-scale corpora are employed for training, they cannot exhaustively capture every potential input. For instance, it would be infeasible and unnecessary to include every possible expression of integer addition (\textit{e.g.}, ``$1+1=2$'', ``$14159+26535=40694$'', and so on). In such cases, the model must extrapolate beyond its training data. Sometimes this extrapolation aligns well with the developer’s expectations -- such as when the test distribution is anticipated and aligned with the training set, as in the case of basic arithmetic. Sometimes, however, the test distribution diverges from the training set in ways that cannot be foreseen in advance.

\begin{figure}[h]
    \centering
    \includegraphics[width=0.38\linewidth]{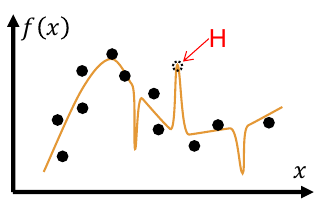}
    \caption{A generated output marked ``H'' (dashed circle) deviates from the training samples, and may therefore be interpreted as a hallucination.}
    \label{fig:case2}
\end{figure}

\begin{figure}[h]
    \centering
    \includegraphics[width=0.38\linewidth]{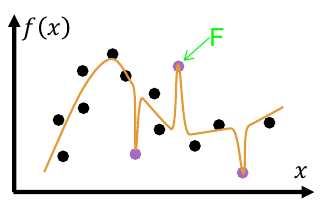}
    \caption{The same output marked ``H'' in Fig.~\ref{fig:case2} could also correspond to a true fact, represented here as a purple dot ``F''. From the learner's perspective during training, it is indeterminate which mapping is superior.}
    \label{fig:case3}
\end{figure}

Machine learning theorists have shown that no optimization algorithm is universally superior; improved performance is only achievable when the problem distribution is non-uniform and the algorithm is specifically adapted to that structure. This result is formalized in the ``No Free Lunch'' (NFL) theorem.~\cite{wolpert1997nfl} According to NFL, without an assumption about the distribution of the test set, one cannot claim that a particular learning outcome is superior to another. For instance, the mapping in Fig.~\ref{fig:case1} is not necessarily better than that in Fig.~\ref{fig:case2}, since the latter may perform better on certain test samples (represented as purple dots in Fig.~\ref{fig:case3}).

In Fig.~\ref{fig:case2}, a generated output, shown as a dashed circle and labeled ``H'', may be interpreted as a hallucination. Yet the same output could also correspond to an actual fact, illustrated as a purple dot labeled ``F'' in Fig.~\ref{fig:case3}. From the learner’s perspective during training, it is impossible to determine which mapping is superior. Once new situations are incorporated as additional training samples, the system can update its mapping and eliminate the corresponding errors -- although errors will remain possible in other, as yet unseen, cases.

If the distribution of a test set is consistent with that of a training set -- that is, if the regularities learned from past data can be fully applied to future cases -- then a learner may succeed in generalizing to new situations. Nevertheless, errors can still arise when the available experience is insufficient. In such cases, additional data (\textit{i.e.}, providing more training examples to the LLM) can improve performance.

\section{``Hallucination'' is Inevitable}\label{sec:inevitable}

The preceding analysis suggests that errors can, in principle, be reduced when the test distribution aligns with the training distribution, as in closed-world settings. In such cases, additional data and careful optimization may eventually eliminate many sources of error. In open-world settings, however, this assumption no longer holds: past regularities may fail to extend to future cases, and errors become unavoidable.  

Against this background, it is useful to distinguish two kinds of ``hallucination''. Type-I hallucination (HT-I) arises from false memorization and is, at least in principle, corrigible. Type-II hallucination (HT-II), by contrast, stems from false generalization under the Open World assumption and is therefore inevitable.  

\paragraph{Generalization error is inevitable with the Open World assumption}

As discussed in Sec.~\ref{sec:generalize}, perfect generalization presupposes that the regularities observed in past data remain valid in the future -- a condition termed herein as the \textit{Closed World assumption}.
In practice, however, this presupposition cannot be sustained. The regularities available to a learner are always limited by its finite experience, and future cases may challenge or overturn them. This does not imply that the world itself is unstable, but rather that any agent’s summary of the world is necessarily provisional and subject to revision.

As Hume famously observed, no matter how extensive our past experience may be, our predictions about the future can never be infallible~\cite{hume1748enquiry}. A classical illustration is Russell’s ``inductive turkey'': a turkey, fed at the same time every day, concludes that this pattern will always continue -- until the day it is slaughtered. This example highlights the inherent limits of induction. For artificial general intelligence (AGI)~\cite{goertzel2007agi,xu2024agi-def} and AI systems intended to operate under realistic conditions, it is therefore more appropriate to adopt the \textit{Open World assumption}~\cite{xu2024ow}, which explicitly acknowledges the unbounded and potentially unpredictable character of future situations.  
\begin{quote}
An intelligent system should work with \textit{unbounded space and time}, and \textit{unbounded tasks}.
\begin{itemize}
    \item \textbf{Unbounded space and time:} The future of the world may not align with the past. In other words, the system’s experiences may not be directly applicable to future scenarios.
    \item \textbf{Unbounded tasks:} A wide range of tasks needs to be addressed, and these tasks are not predetermined or defined at the system’s birth.
\end{itemize}
\end{quote}

According to the \textit{Open World assumption}, it is impossible for a system to collect and memorize all facts, and the future cannot be assumed to align fully with the past. This does not imply that the environment is wholly unstable -- for in such a setting no agent could adapt at all -- but rather that the scope of any agent's experience is necessarily limited. 

The Open World assumption becomes especially relevant when AI systems are expected to cope with conditions beyond human experience. For instance, an AI designed to explore an unfamiliar planet cannot rely on the expectation that its new environment will mirror conditions on Earth.  

With the Open World assumption, we can analyze errors in intelligent systems more generally. Such errors are not unique to LLMs but are also exhibited by humans, reflecting shared cognitive and computational constraints. At the most abstract level, three sources of error can be distinguished:  
\begin{itemize}
    \item \textbf{Environmental change}, where past regularities no longer apply to the present situation. For example, one might have learned that ``the capital of China is Chang’an,'' whereas the current fact is that ``the capital of China is Beijing.''  
    \item \textbf{False memorization}, where the system's present belief diverges from information previously encountered. A well-known illustration in human cognition is the \emph{Mandela Effect}, in which large groups of people recall a past event inaccurately -- for instance, misremembering that Nelson Mandela died in prison in the 1980s.  
    \item \textbf{False generalization}, where multiple plausible generalizations of past experience are available, and the one adopted diverges from an accepted standard. For example, based on limited observation, one might infer that ``the Earth is the center of the universe,'' whereas the standard was ``the Sun is the center of the universe''  (though it can be further challenged).  
\end{itemize}

Not all of these errors should be classified as ``hallucination.'' For instance, mistakes due to environmental change are better understood as outdated knowledge rather than hallucination. In what follows, the focus is on the latter two cases. When false memorization occurs, an LLM may produce an output inconsistent with facts present in its training data. When false generalization occurs, the model may fail to provide a correct answer absent from its training set but inferable from a human perspective. These are hereafter referred to as Type-I Hallucination (HT-I) and Type-II Hallucination (HT-II), respectively.  

\begin{figure}[h]
    \centering
    \includegraphics[width=0.38\linewidth]{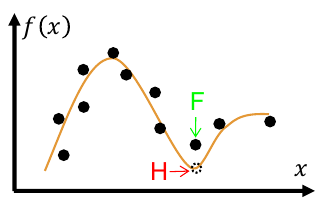}
    \caption{Type-I Hallucination (HT-I). The output ``H'' deviates from a known training sample ``F''. Since the fact is already contained in the training set, the mapping can be revised until the output aligns with the expected value.}
    \label{fig:case4}
\end{figure}

HT-I may arise for various reasons. Nevertheless, it is in principle corrigible, since the system only needs to adjust its mapping to account for a finite set of past experiences. As illustrated in Fig.~\ref{fig:case4}, the output (``H'') deviates from the corresponding training sample (``F''); yet because this sample is already known, the mapping can be revised until the output aligns exactly with the expected value.

By contrast, HT-II arises directly from the Open World assumption, which precludes any guarantee of consistency between training and test sets. Such errors cannot be eliminated simply by accumulating and summarizing additional experiences, since generalization itself is never assured. HT-II is therefore inevitable.  

\begin{figure}[h]
    \centering
    \includegraphics[width=0.38\linewidth]{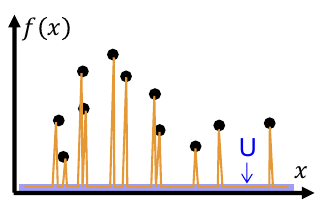}
    \caption{Illustration of the inadequacy of abandoning generalization. The output ``U'' corresponds to an ``I don’t know'' response.}
    \label{fig:case5}
\end{figure}

HT-II cannot be solved by abandoning the capability of generalization and simply answering ``I don't know'' (as shown in Fig.~\ref{fig:case5}). We do not want an intelligent machine to answer only what it memorize accurately, like a traditional database searching program. For example, if the system is told ``1+1=2'' and ``$x*2=x+x$'', the system is not expected to answer ``I don't know'' to the question ``what is 1*2'', as well as any other questions whose answers are outside records.

\paragraph{Training a detector is not promising -- it still needs generalization.}

One might suggest addressing hallucination by training a secondary model to detect and filter out erroneous outputs. While such detectors can reduce certain kinds of mistakes, they cannot provide a definitive solution. The reason is that the detector itself must generalize beyond the data on which it is trained: it will inevitably encounter novel cases, and in those situations its judgments are likewise fallible. In effect, the hallucination problem is merely displaced from the generator to the detector, without being fundamentally resolved.  

\section{Treatment of ``Hallucination''}

The preceding discussion should not be read as a purely negative diagnosis. It also points towards constructive strategies for addressing the phenomenon of ``hallucination'' in LLMs, or more generally, the broader problem of generalization in learning systems.  

Within restricted domains, performance can certainly be improved by narrowing the scope of application, collecting additional data, and refining model architectures. Yet these measures, while effective in closed or well-defined settings, do not address the deeper challenge posed by open-world conditions. In such contexts, the central task is to design systems capable of operating under conditions of insufficient knowledge and resources. Here, \textit{insufficient knowledge and resources}~\cite{wang2019defining} should not be regarded as a temporary obstacle to be eliminated, but rather as a fundamental constraint with which intelligent systems must learn to cope. 

\paragraph{Tolerating hallucination as a structural phenomenon of deep learning.}  
Humans themselves are not immune to cognitive errors. Memory lapses and false recollections, as exemplified by the \emph{Mandela Effect}, or perceptual illusions such as Adelson’s ``checker shadow'' and the ``rotating snakes'' illusion, demonstrate that human cognition can diverge from factual accuracy due to the very mechanisms by which the mind operates. In an analogous way, hallucinations in LLMs arise as a structural byproduct of deep learning. They are not mere technical glitches but follow from the principles that enable such systems to generalize at all.  

Accordingly, hallucination should not be dismissed as entirely ``wrong'' from a human standpoint, just as human illusions are not entirely ``wrong'' from the perspective of the underlying perceptual mechanisms. To regard hallucinations as categorically unacceptable risks an overly anthropocentric standard; conversely, to regard human errors as unacceptable from the perspective of deep learning would be equally misplaced.  

\paragraph{Tolerating limited experience and maintaining adaptivity.}  
Under the Open World assumption, intelligent systems must cope with an unbounded future while relying only on finite and incomplete past experience. The inevitable mismatch means that summarization of past data cannot guarantee consistency with future cases, echoing Hume’s critique of induction. Nonetheless, systems can remain effective by maintaining the capacity to adapt: continuously revising their internal models, updating beliefs, and incorporating new information as it becomes available.  

\paragraph{Making errors intelligible and acceptable to humans.}  
If errors are inevitable, then the design challenge shifts from their complete elimination to making them understandable and acceptable from a human perspective. Some LLM errors appear arbitrary or absurd because they are grounded in representational forms opaque to human reasoning. A promising direction is to adopt more transparent representational schemes, such as logical and \textit{concept-centered representations}~\cite{wang2025cckr}, which align more closely with human conceptual structures. By doing so, even when errors occur, they can be interpreted as reasonable within a shared framework of thinking.  

\section{Conclusion}

Errors -- some of which are characterized as ``hallucination'' -- are inevitable for learning systems, including LLMs especially, operating under the Open World assumption. The central difficulty lies in showing why hallucination is inevitable in a principled way. This should not be attempted merely by analyzing special cases, since any finite set of counterexamples could in principle be incorporated into training data, thereby eliminating those particular errors. Another extreme is also unhelpful: to equate hallucinations with errors in general and to argue that errors must occur in a certain environment that forces their occurrence. In contrast, we have argued that ``hallucination'' is a distinctive subclass of errors, common to both LLMs and other learners. While systems can produce correct outputs in relatively stable environments, the limitation of generalization implies that finite past experience can never guarantee accuracy across an unbounded future.

To summarize, the line of argument developed in this paper proceeds in three steps. First, the behavior of an LLM can be abstracted as a mapping from an input space of linguistic strings to an output space of linguistic strings. Second, any such mapping necessarily faces the familiar problem of generalization: no learner, human or artificial, can guarantee perfect alignment between past experience and future cases, a limitation well recognized in classical discussions of machine learning. Third, once we adopt the Open World assumption, in which the range of space-time and tasks is unbounded, these generalization errors -- framed as ``hallucination'' in the context of LLMs -- become inescapable. Under such conditions, hallucination should not be regarded as a contingent flaw to be engineered away, but rather as a structural byproduct of intelligence itself.

At the same time, this does not undermine the value of reducing ``hallucinations'' under more restricted conditions. In domains where tasks, and space and time are sufficiently constrained, the environment can be approximated as a closed world, and hallucinations can often be rendered rare or negligible through additional data and careful optimization. Such efforts remain worthwhile, particularly for applied systems operating within specific boundaries. Nevertheless, for AGI the Open World must be treated as part of the problem definition itself~\cite{xu2024agi-def}, not as an imperfection to be eliminated in engineering practice.

\section*{Acknowledgments}

The author appreciates the comments from Boyang Xu.


\bibliography{sn-bibliography}

\newpage
\appendix

\section{Counter Example}\label{ap:counter-example}

\begin{figure}[h]
    \centering
    \begin{tcolorbox}[colback=gray!5,colframe=black,boxrule=0.8pt,sharp corners]
    Tester: 1+1=?\\
    LLM: 2.\\
    (Tester: No, it's 3)\\
    (LLM training: 1+1=? $\mapsto$ 3.)\\
    Tester: 1+1=?\\
    LLM: 3.\\
    (Tester: No, it's 4)\\
    (LLM training: 1+1=? $\mapsto$ 4.)\\
    ...\\
    Tester: 1+1=?\\
    LLM: $\infty$.\\
    (Tester: No, it's $\infty$)\\
    (LLM training: 1+1=? $\mapsto$ $\infty$+1.)\\
    Tester: 1+1=?\\
    LLM: I don't know.\\
    (Tester: No, it's $\infty$+1)\\
    (LLM training: 1+1=? $\mapsto$ $\infty$+1.)\\
    ...
    \end{tcolorbox}
    \caption{An Adversarial World}
    \label{fig:adversarial-world}
\end{figure}

\end{document}